\documentclass[letterpaper, 10 pt, conference]{ieeeconf}  % Comment this line out if you need a4paper

\IEEEoverridecommandlockouts                              % This command is only needed if 
\usepackage{times}

\usepackage{multicol}
\usepackage{amsmath,amsfonts,amssymb}
\usepackage{bm}
\usepackage{threeparttable}
\usepackage{algorithmic}
\usepackage{algorithm}
\usepackage{array}
\usepackage[caption=false,font=normalsize,labelfont=sf,textfont=sf]{subfig}
\usepackage{textcomp}
\usepackage{stfloats}
\usepackage{url}
\usepackage{verbatim}
\usepackage{graphicx}
\usepackage{tikz}
\usepackage{etoolbox}
\usepackage{booktabs}
\usepackage{comment}
\usepackage[bookmarks=true]{hyperref}
\usepackage{booktabs} % for \toprule, \midrule, etc.
\usepackage{pifont}   % for \ding{51} and \ding{55}
\usepackage{makecell} % optional: for multi-line headers if needed
\usepackage{colortbl}
\newcommand{\cmark}{\ding{51}} % checkmark
\newcommand{\xmark}{\ding{55}} % crossmark

\usepackage{caption}
\newcommand{\insertfig}{
  \begin{center}
    \includegraphics[width=\textwidth]{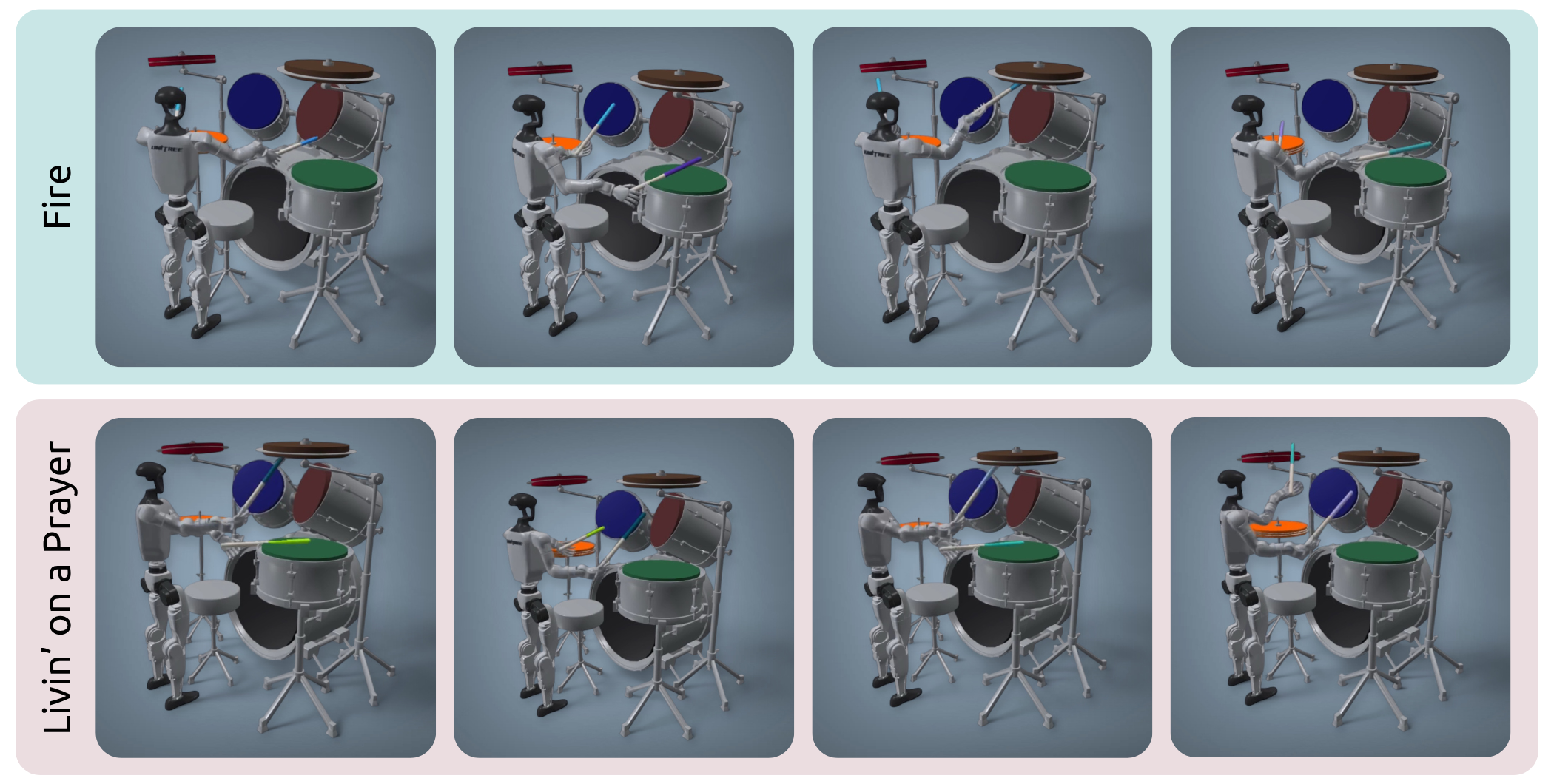}
    \captionof{figure}{Robot Drummer empowers humanoids to learn rhythmic drumming skills across different songs. Top: the robot begins with a tom (reddish brown) hit, then strikes the snare drum (green), cymbal (dark brown), and hi‑hat (orange); Bottom: the robot executes two cross‑over moves, striking cymbal with the right hand and snare with the left, followed by a strike on the second cymbal (dark red).}
    \label{fig:widefig}
  \end{center}
}
\makeatletter
\apptocmd{\@maketitle}{\centering\insertfig\vspace{-1.5em}\setcounter{figure}{0}}{}{}

\makeatother
\title{\LARGE \bf
Robot Drummer: Learning Rhythmic Skills for Humanoid Drumming}

\author{Asad Ali Shahid$^{1,2}$, Francesco Braghin$^{2}$, and Loris Roveda$^{1,2}$ % <-this % stops a space
\thanks{$^{1}$Asad Ali Shahid, and Loris Roveda are with IDSIA, SUPSI, Lugano, Switzerland. 
{\tt\small asadali.shahid@idsia.ch}}%
\thanks{$^{2}$ Asad Ali Shahid, Francesco Braghin, and Loris Roveda are with the Department of Mechanical Engineering, Politecnico di Milano, Italy.}%
}

\begin{document}

\maketitle
\thispagestyle{empty}
\pagestyle{empty}

%%%%%%%%%%%%%%%%%%%%%%%%%%%%%%%%%%%%%%%%%%%%%%%%%%%%%%%%%%%%%%%%%%%%%%%%%%%%%%%%
\begin{abstract}
Musical tasks, like drumming, present unique challenges such as split-second timing, rapid contacts, and multi-limb coordination over performances lasting minutes. In this paper, we introduce Robot Drummer, a simulation framework for humanoid drumming across a diverse repertoire of songs.
We formulate humanoid drumming as the realization of timed contact events encoded as a \textit{Rhythmic Contact Chain}.
To handle the long-horizon nature of musical performance, we decompose each track into fixed-length segments and train a single policy across all segments in parallel using reinforcement learning. Through extensive experiments on over thirty popular tracks, our results demonstrate that Robot Drummer consistently achieves high F1 scores and enables efficient learning of long-horizon musical performances. The learned behaviors exhibit emergent human-like drumming strategies, such as cross-arm strikes, and adaptive stick assignments, demonstrating the potential of reinforcement learning to bring humanoid robots into the domain of creative musical performance. Project page: \href{https://robotdrummer.github.io}{robotdrummer.github.io}
\end{abstract}

\stepcounter{figure} % this increments the figure number by one
%%%%%%%%%%%%%%%%%%%%%%%%%%%%%%%%%%%%%%%%%%%%%%%%%%%%%%%%%%%%%%%%%%%%%%%%%%%%%%%%
\section{Introduction}
Enabling robots to perform music poses significant challenges in dexterity and long-horizon rhythmic control. Drumming, in particular, is a highly complex task that demands precise timing, high-frequency intermittent contacts, dynamic motor coordination, and fine-grained control over acoustic outcome. While robotic manipulation has seen impressive advancements in recent years \cite{fu2024humanplus}, these are often confined to goal-driven tasks with well-defined end states. In contrast, playing music is inherently process-driven, involving intricate timing, and the ability to adapt to changing rhythms. Success depends not on reaching a final configuration, but on sustaining an evolving sequence over a long temporal horizon. Even minor deviations in timing can profoundly affect the emotional quality and impact of a performance. 

Reinforcement Learning (RL) has demonstrated remarkable success in enabling humanoid robots to master goal-driven behaviors, such as getting up \cite{huang2025host}, stabilizing \cite{zhang2025hub}, and tracking agile motions \cite{fu2024humanplus, he2025asap}. In the domain of robotic musicianship, RL has been applied to piano playing with anthropomorphic dexterous hands \cite{robopianist2023}, as well as to a simplified two-DoF underactuated drumming arm \cite{karbasi2024embodied}. Despite these recent efforts, the challenges posed by process-driven tasks like music remain largely unexplored by policy-learning methods. To our knowledge, no prior work has investigated learning-based control for coordinated humanoid drumming.  

In this work we introduce Robot Drummer, a general RL framework that enables a humanoid to perform drum tracks from MIDI (Musical Instrument Digital Interface) transcriptions. Our approach (i) formalizes drumming as a sequence of timed contact events (i.e., the  \emph{Rhythmic Contact Chain}), (ii) makes long performances tractable by training a single policy across fixed-length temporal segments, and (iii) uses dense contact-based rewards that penalize missed and incorrect strikes while rewarding correct hits. Together, these contributions enable policies that achieve high rhythmic fidelity across a diverse set of songs and exhibit emergent, human-like drumming strategies.

\section{Related Works}
\subsection{Learning for Humanoid Control}
Recent humanoid controllers largely follow a motion tracking-based learning paradigm, where policies are trained to reproduce re-targeted reference trajectories. These are typically drawn from motion capture datasets \cite{cheng2024express} or videos \cite{allshire2025visual, zhang2025hub, he2025asap}, often augmented by motion priors \cite{lu2024mobile} and trained with curriculum schedules \cite{huang2025host, zhang2025hub} to enhance robustness and diversity. To tackle contact-rich and dynamic tasks such as parkour jumping, a sequential contact decomposition approach has been used in \cite{zhangwococo} that uses staged reference contacts to simplify reward design. Sim-to-real alignment efforts, such as those in \cite{he2025asap}, have improved fidelity for agile whole-body motion tracking policies in real environments. 

While imitation-based approaches have shown impressive performance in acquiring loco-manipulation skills, they are tied to demonstrated motions in the datasets and optimize primarily over a few-second horizon. For drumming, persistent occlusions by the drum kit, high-speed stick impacts, and dense environment contacts make it difficult to obtain reliable human motion references, whether via motion capture or video-based methods. Moreover, drumming is a contact-rich, long-horizon task, where coordination skills are tightly coupled to temporal contact sequences, that are not well captured by frame-wise motion tracking objectives.

\subsection{Learning to Play Music}
Robotic musicianship has advanced from purely trajectory-based systems \cite{weinberg2007design} to data-driven and learning-based paradigms \cite{robopianist2023, karbasi2024embodied}. Early works used simple arms and rules-based improvisation to interactively play alongside humans \cite{weinberg2007interactive}. More recently, RL has enabled end‐to‐end training of anthropomorphic hands to play piano pieces \cite{robopianist2023}. In drumming, a two-DoF underactuated arm was trained to generate rhythmic patterns under simplified kinematics \cite{karbasi2024embodied}. Additionally, wearable robotic prosthetics have been developed to augment human drumming abilities, using muscle signals to adapt to the musical contexts \cite{yang2021drumming}.

All of the aforementioned works either focus on simplified, low~DOF embodiments for drumming or on floating-hand setups for piano playing. In contrast, Robot Drummer treats percussion as a long-horizon skill that requires coordinated, high-DoF humanoid control and provides a general simulation framework to advance robotic musicianship.

\begin{figure}[b]
    \centering 
    \includegraphics[width=0.95\columnwidth]{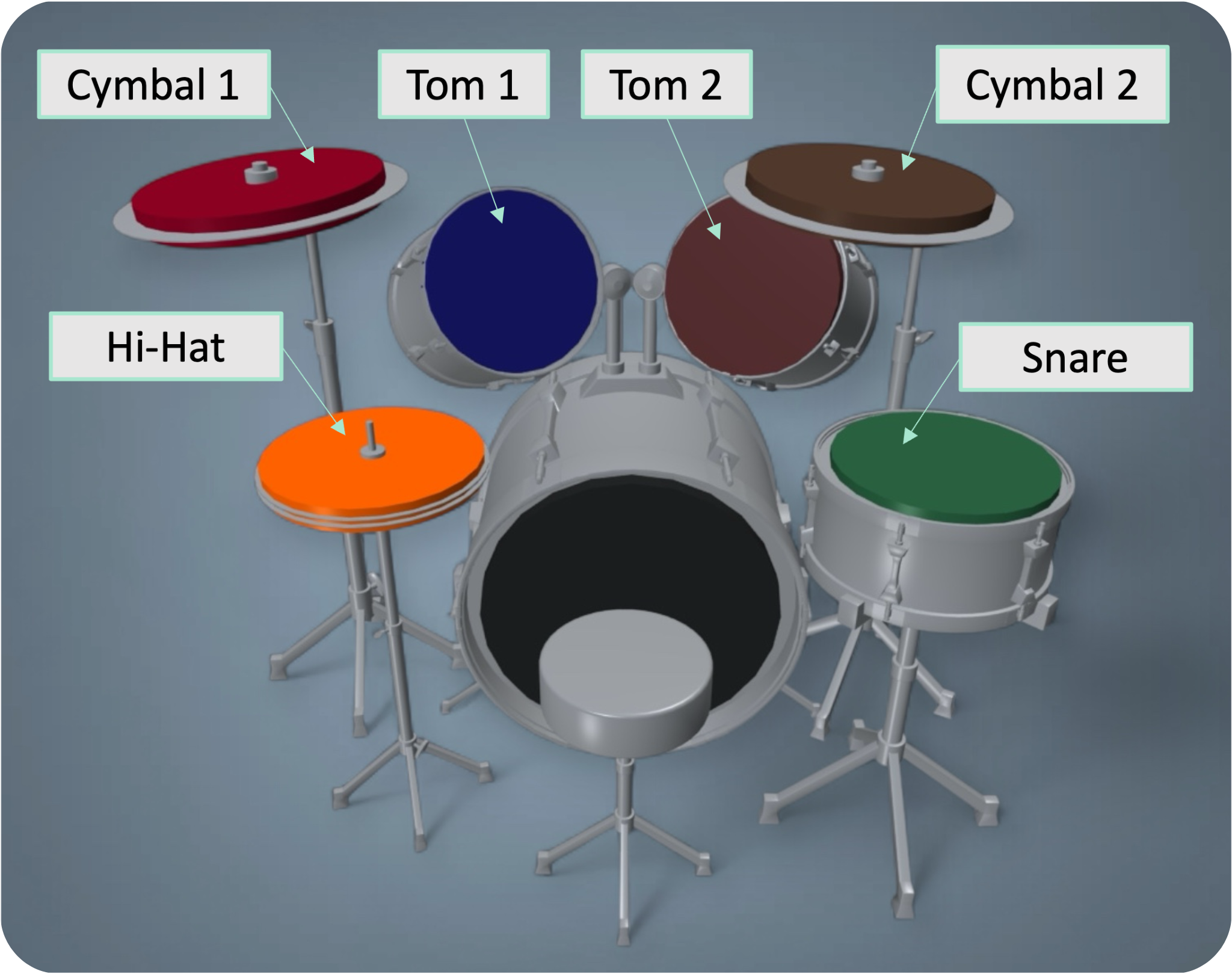}
    \caption{Drum kit configuration used for training and evaluating the humanoid robot drummer (based on \cite{barsky2019multisensory}). The setup includes a snare drum, two toms, a hi-hat, and two cymbals. Each drum is assigned a unique MIDI note, establishing a fixed note-to-drum mapping.}
    \label{fig:drumkit}
\end{figure}

\begin{figure*}[htbp]
    \centering 
    \includegraphics[width=0.99\textwidth]{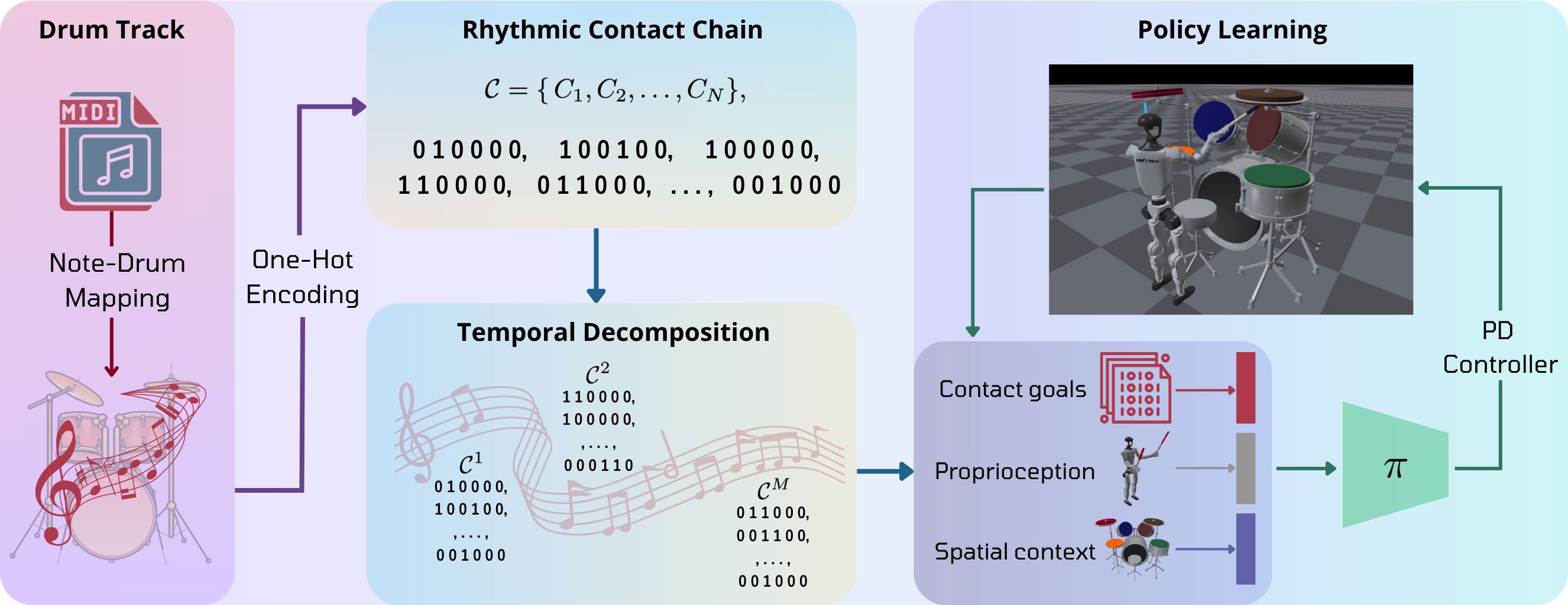}
    \caption{Overview of the Robot Drummer: Starting from a raw MIDI drum track, each note‐onset is first mapped to a specific drum and then converted into a sequence of timed contact events, forming a Rhythmic Contact Chain (RCC). Each event specifies which drum should be struck and when. The RCC is split into fixed-length segments for parallel training. At each timestep, the policy observes the next $L$ contact goals together with robot proprioception and spatial context, and outputs joint position targets for a PD controller.}
    \label{fig:main}
\end{figure*}

\section{Robot Drummer: Learning Rhythmic Skills}
Our goal is to enable a humanoid robot to perform a musical repertoire by striking the correct sequence of drums (as illustrated in Figure~\ref{fig:drumkit}), based on MIDI transcription of sheet music. Each drum is modeled as a collision surface which registers an impact when struck with a sufficient force, causing the synthesizer to produce the corresponding sound. This problem is challenging for three reasons. First, the task is \emph{contact-rich}:~each correct outcome depends on precise, intermittent impacts that are sensitive to small spatial and timing errors. Second, the task is \emph{long-horizon}:~songs can last from a few minutes to over six minutes, containing hundreds of hits, that lead to poor exploration. Third, effective drumming couples \emph{temporal} and \emph{spatial} planning:~choices about which stick to use and where to position the hands affect the ability to execute upcoming contacts. To address these issues we represent a drum score as a \emph{Rhythmic Contact Chain} (RCC) (Section~\ref{sec:rcc}), split each RCC into fixed-length segments that are trained in parallel (Section~\ref{sec:decompose}), and design task-agnostic contact rewards that provide graded signals for correct, wrong, and missed strikes (Section~\ref{sec:policy}). An overview of our approach is illustrated in Figure~\ref{fig:main}.

\subsection{Music Representation}
\label{sec:music-rep}
While music can be represented as raw waveforms or spectrograms, these formats present challenges for robots, particularly when processing continuous data and designing appropriate reward functions. We use MIDI as a musical representation, which provides a standardized way to encode instruments' timing in digital music. Specifically, we use publicly available community-created MIDI files, which are individual recreations of popular songs that may differ from original recordings in timing and instrument assignment. From a full MIDI sequence, we extract only the drum events, each containing the note, start time, and end time. To relate these events to a drum kit, we map distinct MIDI notes to corresponding drums (see Figure \ref{fig:drumkit}). This mapping is song-specific, as the same drum may correspond to different notes across songs. In addition, each drum can have multiple articulations, which are variations in how the drum is played, for example, open or closed hi-hat. These articulations produce distinct sounds and are represented as separate note numbers on a General MIDI percussion map \footnote{\href{https://musescore.org/sites/musescore.org/files/General MIDI Standard Percussion Set Key Map.pdf}{MIDI Percussion Map}}. When multiple articulations appear for the same drum in a song, we select only the most frequent one to simplify the representation. Extending our framework to multiple articulations is left for future work. 

\subsection{Rhythmic Contact Chain}
\label{sec:rcc}
Humanoid drumming performance can be viewed as a sequence of precise and coordinated contact events, where each event corresponds to drum(s) strikes at a specific time. We model this sequence as the \textit{Rhythmic Contact Chain}, which represents the timed contacts the robot must execute to perform a musical piece. Formally, we define it as:

\begin{equation}
    \mathcal{C} =  \{\,C_1, C_2, \dots, C_N\},
\end{equation}

where $C_i$ denotes the contact step at time $t_i$ and may include multiple contact pairs corresponding to the target drum strikes. Each contact step is defined as:

\begin{equation}
     C_i = \bigl(D_i,\,t_i,\,V_i,\,S_i\bigr),
\end{equation}

with
\[
  D_i \subseteq \mathcal{D},\qquad
  V_i \in \mathbb{R}^{|\mathcal{D}|},\qquad
  S_i \subseteq \{s_{\text{L}}, s_{\text{R}}\},
\]

where $D_i$ denotes the set of drums to strike at time $t_i$, as specified by the drum track, $V_i$ is a per-drum velocity vector aligned with the drum set $\mathcal{D}$ (i.e., $(V_i)_d$ gives the MIDI velocity for drum $d$ when $d\in D_i$), $S_i$ optionally specifies which stick(s), left or right, may be used to execute the contacts, and $\mathcal{D}$ contains all the drums in a drum kit as shown in Figure \ref{fig:drumkit}. In this work we do not condition the policy or rewards on $V_i$, treating all strikes as equal intensity, and do not provide explicit assignments $S_i$, letting the policy discover stick-to-drum coordination autonomously through contact rewards in Section~\ref{sec:rew}. Both terms are retained to support extensions to expressive dynamics or explicit stick assignment.

\subsection{Temporal Decomposition}
\label{sec:decompose}
Performing an entire drum track (several minutes long) in a single RL episode leads to sparse rewards and poor exploration, since thousands of timesteps must be rolled out sequentially before any reset, limiting parallelism and delaying exposure to later song sections. To address this, we introduce a temporal decomposition approach that partitions the full drum track into a sequence of fixed-length training segments, all trained in parallel. Each segment corresponds to a subset of the contact steps in RCC. Specifically, we divide $\mathcal{C}$ into $M$ segments:

\begin{equation}
\mathcal{C}^{(m)}
\;=\;
\bigl\{\,C_i : (m-1)\,P + 1 \le i \le m\,P \bigr\},
\end{equation}

where $P~=~N/M$ is the number of contact steps per segment, and $m = 1,2,\dots,M$ represents the segments of a track. Each segment $\mathcal{C}^{(m)}$ defines a temporally localized portion of the song. This decomposition transforms a single long-horizon task into multiple shorter training segments, significantly improving the sample efficiency of learning rhythmic skills. 

\subsection{Policy Training}
\label{sec:policy}
We formulate humanoid drumming as a finite-horizon extended Markov Decision Process (MDP), defined by a tuple $(\mathcal{S, A, T, R, \gamma, C})$, where $\mathcal{S}$ is the state space, $\mathcal{A}$ is the action space, $\mathcal T$ governs the transition dynamics, $\mathcal T:\mathcal S \times \mathcal A \rightarrow \mathbb P(\mathcal S)$, $\mathcal R:\mathcal S \times \mathcal A \rightarrow \mathbb R$ is the reward function, $\gamma \in [0, 1]$ represents the discount factor, and $\mathcal C$ is the contact chain defined in Section~\ref{sec:rcc}. The objective is to learn an optimal policy $\pi_\theta$ that maximizes the expected cumulative reward $\mathbb{E}_{\pi_{\theta}}\left[\sum_{t=0}^{H-1} \gamma^t r_t \right] $ over the horizon $H$. To optimize the policy, we employ Proximal Policy Optimization (PPO) \cite{schulman2017proximal}, an on-policy RL algorithm known for its stability and efficiency in large-scale parallel training environments.

\subsubsection{States}
The state \( \bm{s}_t \) is composed of three components: the robot's proprioception \( \bm{s}_t^{\mathrm{p}} \), spatial information related to the drums and sticks \( \bm{s}_t^{\mathrm{sp}} \), and contact goals \( \bm{s}_t^{\mathrm{c}} \). The proprioceptive state is defined as
\(
\bm{s}_t^{\mathrm{p}}~\triangleq~[\,\bm{q}_t,\; \dot{\bm{q}}_t,\; \bm{a}_{t-1}\,],
\)
where \( \bm{q}_t \in \mathbb{R}^{15} \) are the joint positions, \( \dot{\bm{q}}_t \in \mathbb{R}^{15} \) are the joint velocities, and \( \bm{a}_{t-1} \in \mathbb{R}^{15} \) is the action applied at the previous timestep. The spatial context \(
\bm{s}_t^{\mathrm{sp}} \in \mathbb{R}^{24}\) encodes the positions of the drums and sticks. The contact goals state is defined as
\(
\bm{s}_t^{\mathrm{c}}~\triangleq~[\,\bm{c}_t, \bm{c}_{t+1}, \dots, \bm{c}_{t+L}\,],
\)
where \( \bm{c}_{t+i} \in \mathbb{R}^{6} \) is a binary encoding of target drum strikes at timestep \(t+i\) and \(L\) is the lookahead horizon over future steps to provide rhythmic context. The lookahead horizon controls how far into the RCC the robot can see. The contact goals state is akin to a time-phase variable in motion tracking policy learning \cite{peng2018deepmimic} and effectively functions as a discrete phase marker, informing the policy which drum(s) it should be targeting at that moment in the musical piece. We evaluate the impact of contact goals in our ablation study by comparing it to a continuous time phase variable.

\subsubsection{Actions}
The policy's actions \( \bm{a}_t \in \mathbb{R}^{15}\) correspond to the target joint angles for a Proportional-Derivative (PD) controller, which applies torques to the robot's joints. The torque at timestep \(t\) is computed as:

\begin{equation}
\bm{\tau}_t~=~ \bm{K}_p( \bm{q}_{t}^{d} - \bm{q}_{t}) - \bm{K}_d(\bm{\dot{q}_t}),
\end{equation}

where \(\bm{q}_{t}^{d}\) is the desired joint configuration, defined as \(\bm{q}_{t}^{d}~=~\beta\bm{a}_{t} + \bm{q}_{0}\); \(\beta\) is an action scaling factor, \(\bm{q}_0\) represents default joint positions of each joint, and \(\bm{K}_p\) and \(\bm{K}_d\) are the stiffness and damping coefficients of the PD controller, respectively.

\subsubsection{Rewards}
\label{sec:rew}
Effective drumming requires executing hundreds of precisely timed strikes on the correct drums while avoiding any wrong hits. Optimizing solely with the sparse rewards fails to guide the policy toward right rhythmic behaviors.  We hypothesize that every contact event--correct, wrong, or missed--provides crucial information for learning the fine‑grained timing and coordination needed in drumming. We thus propose unified task reward as:

\[
r_t
= r_t^{\mathrm{contact}}
+ r_t^{\mathrm{reg}},
\]
where \(r_t^{\mathrm{contact}}\) provides dense feedback for contact events, and \(r_t^{\mathrm{reg}}\) contains regularization terms. The contact reward is defined as:

\[
r_t^{\mathrm{contact}}
= w_{\mathrm{c}}\,r_t^{\mathrm{correct}}
+ w_{\mathrm{w}}\,r_t^{\mathrm{wrong}}
+ w_{\mathrm{m}}\,r_t^{\mathrm{missed}}
+ w_{\mathrm{p}}\,r_t^{\mathrm{prox}},
\]

where the terms respectively reward correct target strikes, penalize wrong and missed contacts, and encourage proximity to target drums at expected hit times. The first three terms directly correspond to the contact events in \textit{Rhythmic Contact Chain}, defined in Section~\ref{sec:rcc}, while the last term is a proximity reward which encourages the agent for moving sticks closer to target drums at moments when the hits are expected. A full list of reward terms and weights is provided in the supplementary material. The weights are based on an intuition that missing a strike is worse than hitting a wrong drum, while correct hits should be rewarded more to encourage well-timed strikes.

\subsection{Music State Initialization}
State initialization plays a crucial role in training RL policies, where the key idea is to initialize the agent's starting state at random phases along a reference motion trajectory. Recent works have shown the effectiveness of state initialization in tracking reference motions \cite{he2025asap, peng2018deepmimic}. To apply this idea in the context of drumming, we propose \emph{Music State Initialization}, where we randomly sample contact steps within a segment \(\mathcal{C}^{(m)}\) of RCC and initialize the corresponding contact goals state at those steps. This effectively randomizes the starting point of the drum track and exposes the policy to high reward contact states early on in the training.

\begin{figure*}[t]
    \centering
    \begin{tikzpicture}
      \node[anchor=south west, inner sep=0] (img1) at (0,0) {\includegraphics[width=\textwidth]{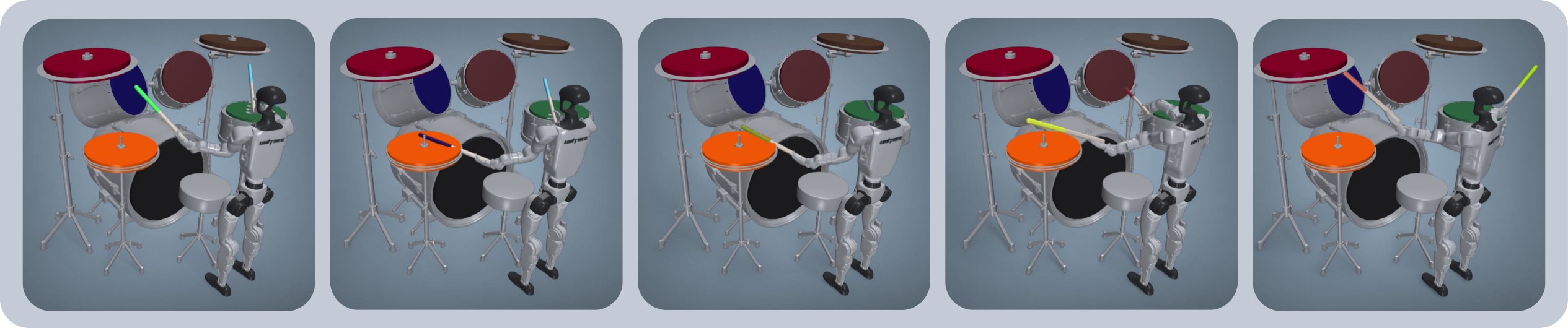}};
    \end{tikzpicture}
    \vspace{0.3em}
    
    \begin{tikzpicture}
      \node[anchor=south west, inner sep=0] (img2) at (0,0) {\includegraphics[width=\textwidth]{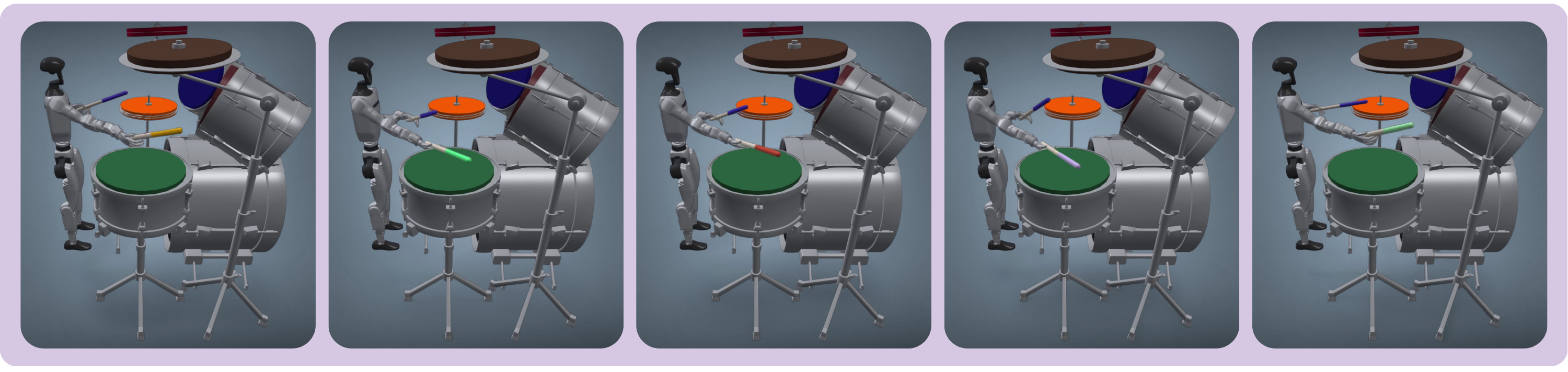}};
      \node[anchor=south west, inner sep=0] at (0,0) {\includegraphics[width=\textwidth]{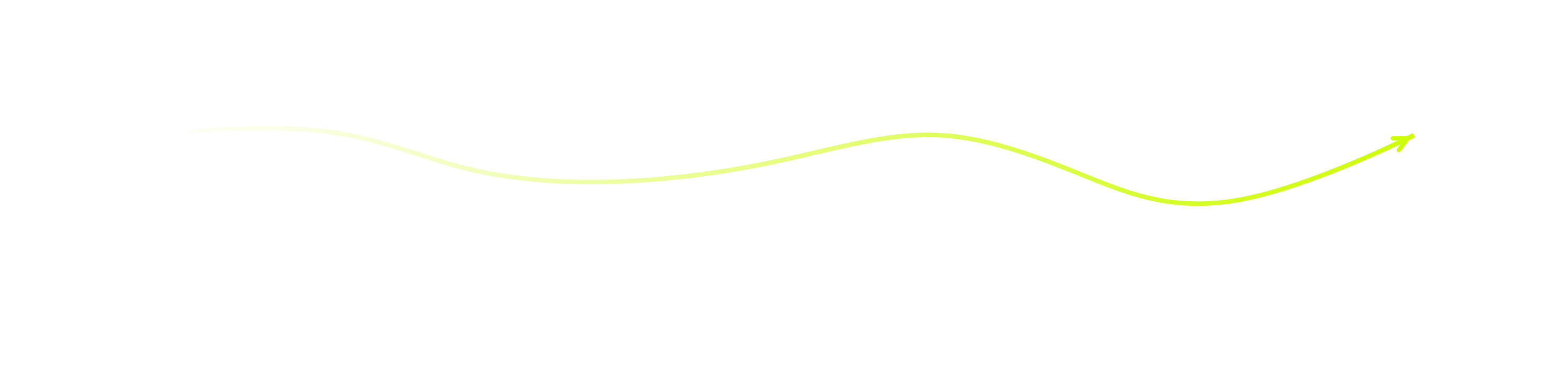}};
    \end{tikzpicture}
    \caption{Qualitative snapshots of challenging drumming sequences. Top: a complex drum-fill segment from ``Linkin Park -- In the End,'' showing the robot executing multiple strikes across four drums in under two seconds. Bottom: a fast drum roll on the snare during ``Nirvana -- Come As You Are'' within a one-second interval, highlighting the precise positioning of the right hand to execute rapid strikes.}
    \label{fig:drum_sequences}
\end{figure*}

\section{Experimental Setup}
We evaluate Robot Drummer through a series of experiments in Isaac Gym simulation environment using a Unitree G1 humanoid robot. The robot is controlled through 15 degrees of freedom (DoF)—7 DOF for each arm and 1 DOF for torso—with drumsticks rigidly mounted to the hands and all striking motions generated via upper body. Our experiments are designed to investigate the following key questions:

\begin{itemize}
\item \textbf{Q1.} How does the Robot Drummer perform across songs of varying complexity?
\item \textbf{Q2.} What role does each component of the Robot Drummer’s design play in its overall performance?
\item \textbf{Q3.} How does a policy trained on multiple songs compare against specialist policies?
\end{itemize}

\subsection{Quantifying Rhythmic and Spatial Complexity}
To address \textbf{Q1}, we first identify the factors that affect the difficulty of drumming in a given song and then train specialist policies. The primary difference in songs complexity lies in the pattern of drum hits, both in time and space. To systematically assess these patterns, we introduce a set of features that capture the song’s key characteristics. These features, described in detail below, provide a quantitative measure of a song's complexity.

\subsubsection{Number of Drums Used}
The number of distinct drums in a song directly affects its rhythmic and spatial complexity. More drums introduce variation into the pattern, requiring greater coordination and contributing to a more intricate rhythm. In this work, the maximum number of drums that can be used in a song is 6, as illustrated in Figure~\ref{fig:drumkit}. 

\subsubsection{Beats per Minute}
The Beats per minute (BPM) measures the tempo or speed of the song. It indicates how fast or slow the song is played, determining the pace at which drumming occurs.

\subsubsection{Normalized Pairwise Variability Index (nPVI)}
The normalized Pairwise Variability Index (nPVI) \cite{toussaint2013pairwise} quantifies the relative variability in timing between successive drum hits, known as inter-onset intervals (IOIs). For each adjacent pair of IOIs, we compute the absolute difference divided by their mean, then average these values across the entire song. A higher nPVI reflects greater temporal irregularity and thus a more challenging rhythm for the robot to execute.

\subsubsection{Hit Distribution Entropy}
Some songs concentrate drum hits on a few drums, while others spread them more evenly. We use Shannon entropy to quantify how dispersed the strikes are. A more even distribution of drum hits yields higher entropy, indicating that the robot must switch between different drums throughout the song and coordinate wider spatial movements between the hits.

\subsubsection{Polyphony}
Although the humanoid robot is physically limited to striking two drums at once (one per stick), MIDI drum sequences often encode complex polyphonic textures with three or more drums at the exact same moment, e.g., simultaneous hit on the snare, tom, and cymbal. We quantify this \textit{polyphony} by measuring the fraction of time‐steps where the MIDI drum track requests more than two simultaneous hits. These overlapping notes signal moments in the song when the robot must prioritize which drums to hit.

\subsection{Evaluation Metrics}
To evaluate the drumming performance of learned policies, we use precision, recall, and F1 score by comparing ground‐truth sequence of desired drum strikes in RCC (see Section~\ref{sec:rcc}) to the robot’s actual hits at each timestep. Precision measures the proportion of the robot’s strikes that correctly align with RCC events (i.e., on the intended drums) over the entire song, while recall captures the fraction of ground-truth contact events the robot actually executes. The F1 score is the harmonic mean of precision and recall and provides a single, balanced metric of overall rhythmic fidelity on a 0-1 scale, with higher values indicating better performance.

\begin{table*}[t]
\centering
\caption{Quantitative evaluation of policies on songs with zero polyphony}
\label{tab:zero_poly}
\resizebox{0.96\textwidth}{!}{
\small
\begin{tabular}{l c c c c c}
\toprule
Song                        & \#Drums & Entropy & nPVI &  BPM  & F$_1$ $\uparrow$ \\
\midrule
David Bowie - Rebel Rebel                 & 2 & 0.27 & 8.76 & 130 & 0.985 $\pm$ 0.008  \\
The White Stripes - Seven Nation Army     & 4 & 0.97 & 6.71 & 120 & 0.977 $\pm$ 0.004  \\
Nirvana - Lithium                         & 6 & 0.74 & 22.10 & 112 & 0.954 $\pm$ 0.018\\
Jimi Hendrix - Fire                       & 6 & 0.67 & 29.48 & 156 & 0.943 $\pm$ 0.021  \\ 
Dave Brubeck - Take Five                  & 4 & 0.71 & 51.86 & 167 & 0.908 $\pm$ 0.034 \\
Linkin Park - In The End                  & 6 & 0.61 & 32.68 & 86 & 0.901  $\pm$ 0.020 \\
Bon Jovi – Livin’ on a Prayer             & 6 & 0.84 & 82.93 & 123 & 0.885  $\pm$ 0.029 \\
The Police - Roxanne                      & 6 & 0.73 & 21.19 & 136 & 0.878  $\pm$ 0.024 \\
\bottomrule
\end{tabular}}
% \begin{tablenotes}

% \footnotesize
% \item[1] All songs have zero time signature changes, except for "Bon Jovi – Livin’ on a Prayer" (4).
% \end{tablenotes}
\end{table*}

\section{Results}
To evaluate the performance of Robot Drummer, we trained over 30 specialist policies on popular rock and metal tracks of varying complexities. Table~\ref{tab:zero_poly} summarizes the complexity metrics for songs with zero polyphony (no more than 2 simultaneous hits) alongside corresponding F1 scores achieved by the policies. The results show that specialist policies consistently achieve high F1 scores. Songs with fewer drums and regular rhythmic patterns (low nPVI) yield near-perfect performance, while increased drum count and higher rhythmic variability (high nPVI) drops the F1 score. Extended results for songs with non-zero polyphony are provided in the supplementary material. Additionally, Figure~\ref{fig:drum_sequences} provides qualitative snapshots of challenging drumming sequences: a drum fill, which involves a rapid succession of hits across multiple drums to transition between sections, and a snare roll, which involves high-speed hits on a snare drum. These examples demonstrate the robot’s ability to perform both spatially and temporally demanding patterns with precise coordination.

To quantify which song characteristics most strongly affect performance, we also computed Spearman’s rank correlation between each feature and per-song F1 score across all specialist policies. We found that rhythmic irregularity (nPVI) shows the strongest negative correlation with F1 score (\(\rho = -0.52\)), indicating that songs with highly uneven timing between hits are most challenging. The number of drums used in a song also correlates negatively with F1 score (\(\rho = -0.42\)). This trend aligns with our expectations as each additional drum target amplifies the difficulty of policy learning due to a larger spatial manifold. The policy needs to  coordinate larger movements across a wider space with a greater potential for incorrect and missed hits.

\begin{figure}[htbp]
  \centering
  \includegraphics[width=\columnwidth]{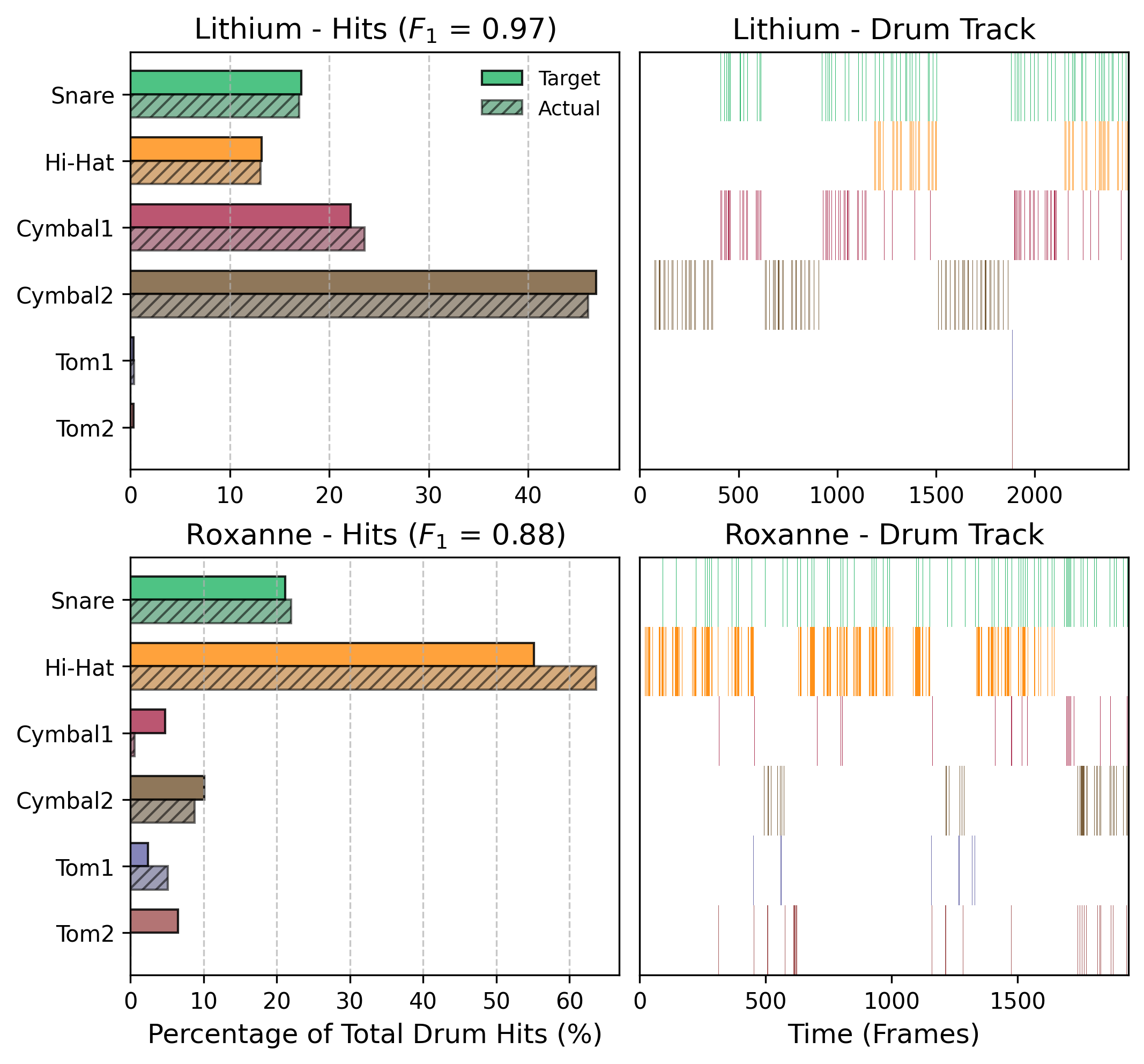}
  \caption{Performance comparison of Nirvana - Lithium (top) and The Police - Roxanne (bottom), showing target drum hits and executed strikes (left) alongside the original drum track (right). Lithium’s regular, well‐spaced rhythm allows the robot to hit nearly every target, whereas Roxanne’s irregular, densely interleaved hits force rapid spatial switches that the policy cannot execute, leading to missed strikes and lower F1 score. Drum tracks are visualized at 10 frames per second~(fps), while the policies operate at 50 fps.}
  \label{fig:comp}
\end{figure}

Figure~\ref{fig:comp} compares the performance of specialist policies on two representative songs (Nirvana - Lithium and The Police - Roxanne), by overlaying the target hits versus actual robot strikes alongside the original drum track. The two policies achieve notably different F1 scores despite similar values for all rhythmic complexity metrics. Lithium's drum pattern is highly regular and evenly spaced, whereas Roxanne’s pattern has irregular cluster of hits with many switches between drums in quick succession (see Figure~\ref{fig:comp} (right)). The regularity in Lithium gives the robot sufficient time to swing each stick between target drums, resulting in an almost perfect F1 score (0.977). By contrast, Roxanne's sequence interleaves distinct drum targets in short time windows.  For example, the transition from cymbal2 to tom2 occurs rapidly, and as a result, the policy stays near cymbal2 and never attempts to hit tom2. Likewise, cymbal1 hits are squeezed into very narrow gaps between hi-hat strikes, and the policy tends to prioritize the more frequent pattern, staying locked in on hi-hat while missing the majority of cymbal1 strikes. These examples reflect how the policies learn a preference for spatially efficient actions under tight timing constraints.

\subsection{Ablation Analysis}
To assess the individual contributions of our design choices, we perform an ablation study of the contact reward components and the observation, reporting mean and standard deviation of F1 scores across five runs in Table~\ref{tab:abl}. In the reward ablation, we retain proximity and regularization terms and compare three variants: (i) correct-strike reward only, (ii) correct plus wrong strikes, and (iii) the full dense contact reward (correct + wrong + missed) strikes. The results demonstrate that the full reward formulation is essential for achieving robust rhythmic accuracy, as omitting either wrong or missed strike terms degrades F1 across five songs. 

We next evaluate the effect of removing key components from the policy's input state during training. We find that the policies trained without the contact goals collapse to near-zero F1, indicating they lack the ability to anticipate upcoming drum strikes. We also examine the role of an explicit time phase signal by replacing the contact goals with a continuous, normalized phase variable \cite{peng2018deepmimic} ranging between~$0-1$. Phase-conditioned policies likewise perform poorly, suggesting that coarse temporal context alone is insufficient to acquire drumming skills. The time phase variable informs the policy "how far through the segment" it is but not "which drum hit is next", highlighting the necessity of explicit goal encoding. Collectively, these results show that humanoid drumming requires both dense contact rewards and structured state representation that encode future rhythmic targets.

Finally, we evaluate the impact of removing temporal decomposition scheme by training policies end-to-end on full-length tracks rather than on fixed-length segments. While the final F1 scores remain comparable in both settings, we observe a substantial increase in wall-clock training time: policies trained without decomposition require approximately 8–9 hours to converge, compared to just 2–3 hours with temporal segmentation on the same hardware. Detailed results are reported in the supplementary material.

\begin{table*}[htbp]
\centering
\caption{Ablation of Reward and Observation Components. F1 scores are reported as mean $\pm$ std over 5 runs.}
\label{tab:abl}
\definecolor{rewardcolor}{RGB}{225, 245, 225}
\definecolor{songcolor}{RGB}{220, 230, 255}
\definecolor{obscolor}{RGB}{235, 235, 235}
\resizebox{\textwidth}{!}{
\begin{tabular}{ccc c c c c c c}
\toprule
\multicolumn{3}{>{\columncolor{rewardcolor}}c}{Reward Components} & 
\multicolumn{6}{>{\columncolor{songcolor}}c}{Song} \\
\midrule
% \multicolumn{9}{>{\columncolor{obscolor}}c}{Reward Ablation} \\
\midrule
Correct Strikes & 
Wrong Strikes & 
Missed Strikes & 
Rebel Rebel &
Lithium & 
Fire & 
In The End & 
Livin’ on a Prayer & 
Roxanne \\
\midrule
\cmark & \xmark & \xmark & 0.64 $\pm$ 0.16 & 0.70 $\pm$ 0.04 & 0.37 $\pm$ 0.06 & 0.60 $\pm$ 0.05 & 0.35 $\pm$ 0.04 & 0.52 $\pm$ 0.04 \\
\cmark & \cmark & \xmark & 0.96 $\pm$ 0.00 & 0.86 $\pm$ 0.03 & 0.91 $\pm$ 0.01 & 0.70 $\pm$ 0.04 & 0.62 $\pm$ 0.16 & 0.69 $\pm$ 0.00 \\
\cmark & \cmark & \cmark & 0.99 $\pm$ 0.00 & 0.95 $\pm$ 0.02 & 0.94 $\pm$ 0.02 & 0.90 $\pm$ 0.02 & 0.89 $\pm$ 0.03 & 0.88 $\pm$ 0.02 \\
\midrule
\multicolumn{3}{>{\columncolor{rewardcolor}}c}{Observation Components} & 
\multicolumn{6}{>{\columncolor{songcolor}}c}{Song} \\
% \multicolumn{9}{>{\columncolor{obscolor}}c}{Observation Ablation} \\
\midrule
Proprioception \& Spatial & 
Phase Variable& 
Contact Goals & 
Rebel Rebel &
Lithium & 
Fire & 
In The End & 
Livin’ on a Prayer & 
Roxanne \\
\midrule
\cmark & \xmark & \xmark & 0.07 $\pm$ 0.03 & 0.07 $\pm$ 0.02 & 0.03 $\pm$ 0.00 & 0.06 $\pm$ 0.01 & 0.01 $\pm$ 0.00 & 0.01 $\pm$ 0.00 \\
\cmark & \cmark & \xmark & 0.12 $\pm$ 0.02 & 0.09 $\pm$ 0.03 & 0.05 $\pm$ 0.00 & 0.10 $\pm$ 0.00 & 0.01 $\pm$ 0.00 & 0.02 $\pm$ 0.00 \\
\cmark & \xmark & \cmark & 0.99 $\pm$ 0.00 & 0.95 $\pm$ 0.02 & 0.94 $\pm$ 0.02 & 0.90 $\pm$ 0.02 & 0.89 $\pm$ 0.03 & 0.88 $\pm$ 0.02 \\
\bottomrule
\end{tabular}}
\end{table*}

\subsection{Multi-Song Policies}
\begin{figure}[htbp]
  \centering
  \includegraphics[width=0.9\columnwidth]{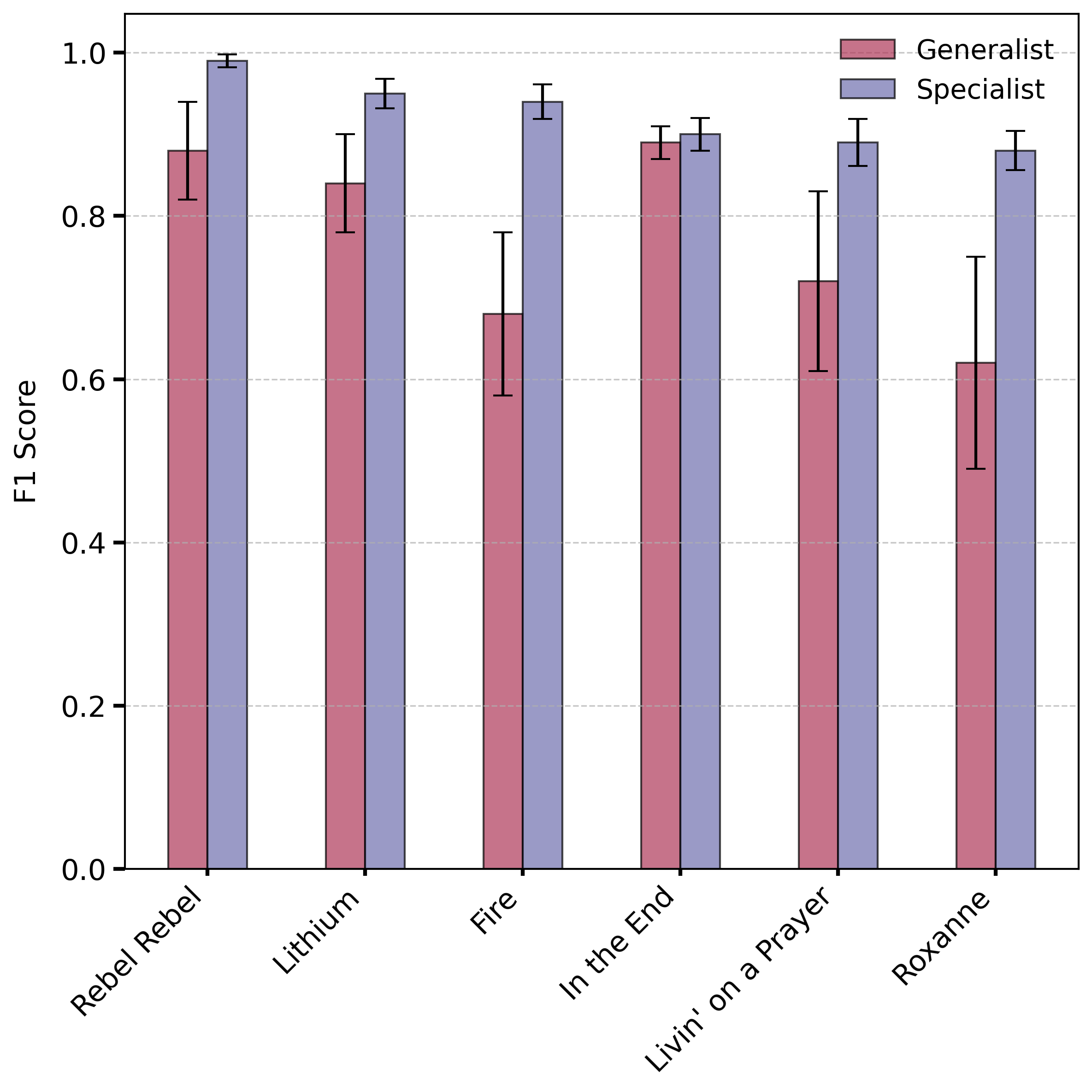}
  \caption{F1 scores for specialist and multi-song policies on selected tracks, illustrating the performance drop when a single policy is trained across multiple rhythms.}
  \label{fig:generalist}
\end{figure}

We also trained policies to play multiple songs with a single controller. A policy trained on six songs from Table~\ref{tab:zero_poly} performed substantially worse on each song than specialist policies (Figure~\ref{fig:generalist}). Songs with higher rhythmic irregularity or faster tempos, such as Livin’ on a Prayer and Fire, exhibited the largest performance drops. In contrast, specialist policies achieved F1 scores near or above 0.9 on these tracks. Training on all eight songs further degraded performance across all songs, indicating that naive multi-song training introduces trade-offs between tasks. Combining multiple rhythmic patterns (e.g. irregular timing of Livin’ on a Prayer and Fire's fast tempo) in one policy produced negative transfer \cite{lan2023contrastive}, reducing task-level performance. Songs differ widely in drum count and rhythmic structure, forcing the policy to accommodate conflicting requirements. As a result, the policy often adopts conservative strategies, focusing on more frequent drum targets while neglecting others.

We also conducted a small listener study with 15 participants to evaluate the perceived musical qualities of the Robot Drummer. Participants rated the performances on timing consistency, expressiveness, naturalness, and overall enjoyment using a 5-point Likert scale. The robot achieved above-average ratings for timing (3.6/5) and expressiveness (3.4/5), lower scores for naturalness (2.5/5), and moderate overall enjoyment (3.1/5). Full study details are reported in the supplementary material.

\section{Discussion}
Our experiments demonstrate that specialist policies can achieve high rhythmic fidelity for humanoid drumming (F1~$\ge$~0.9 on many tracks), and accurately execute complex drum fills and rolls (see Figure~\ref{fig:drum_sequences}). Beyond timing accuracy, the policies exhibit emergent spatial strategies: for example, the robot executed multiple cross-over strikes while performing Livin’ on a Prayer. Such maneuvers were not encoded in the reward; they arose as the policy searched for efficient ways to reach distant drums.
Our analysis further indicates that spatial complexity (number of drums) and temporal variability (nPVI) are strongly correlated with performance degradation. Songs with many quick switches between drums caused the robot to favor the most frequent targets and miss others. In addition, training a single policy across multiple songs led to a consistent drop in F1 scores, highlighting the challenges of multi‑task RL for rhythmic motor control. Together, these findings suggest that effective drumming requires tightly coupling temporal context with spatial planning (how the robot hits one drum directly influences its ability to reach the next targets). Our dense contact rewards and future contact goals encoding implicitly accounted for this. Indeed, we observed policies dynamically reassign sticks based on the musical context. For instance, in Livin’ on a Prayer, the policy typically used the right stick for the snare (a spatially efficient choice) but switched to the left stick when a cymbal2 hit immediately followed the snare hit, ensuring the right hand remained available for the cymbal while still executing the snare strike on time.

\subsection{Limitations}
Our experiments are conducted in simulation and do not model several physical aspects central to real-world drumming, including sound production, stick rebound dynamics, and grip compliance. As a result, the learned policies optimize for contact timing and spatial coordination, but do not reason about acoustic outcomes. We hypothesize that these unmodeled effects can be learned from real data. This points to a broader and relatively unexplored research question: can robots be trained to control the sound they produce? Optimizing for acoustic targets would force policies to implicitly control impact dynamics, closing the loop between motion and sound.

% \subsection{Path toward a Real System}
Transferring the learned policies to a physical system will require (i) a compliant hand–stick interface to better handle contact dynamics, and (ii) a residual control component that applies corrective action adjustments to compensate for unmodeled dynamics. A practical transfer pathway is to use an electronic drum kit that outputs MIDI events, enabling direct comparison between executed and target hits to fine-tune timing and strike intensity, thereby compensating for acoustic mismatch. While full acoustic modeling remains challenging, a sound model learned from real-world data could be integrated into simulation to train sound-aware policies. 

\section{Conclusion}
In this work, we have introduced Robot Drummer, a general RL-based framework that enables a humanoid to perform complex drumming patterns. By modeling drumming as a rhythmic contact chain and leveraging temporal segmentation with dense contact rewards, our approach scales to long musical sequences and achieves high F1 scores on many songs. The learned policies exhibit emergent strategies, such as cross-over strikes and dynamic stick assignment, without explicit encoding. Future work will focus on sim-to-real deployment and adapting policies to match acoustic targets, advancing sound-aware control that could enable new forms of collaboration between robots and human musicians.

\section*{Acknowledgments}
The authors thank Mirko Nava for insightful discussions and Marco Maccarini for setting up the training infrastructure, both at IDSIA.

\bibliographystyle{IEEEtran}
\bibliography{bibliography}

\section*{APPENDIX}
\subsection{Implementation Details}

\subsubsection{Rewards}
Table~\ref{tab:rew} lists both the contact reward terms and regularization terms together with the weights used in all experiments.

\begin{table}[htbp]
\centering
\caption{Reward Components}
\label{tab:rew}
\resizebox{\columnwidth}{!}{ 
\begin{tabular}{lcc}
\toprule
\textbf{Term} & \textbf{Expression} & \textbf{Weight} \\
\midrule
\multicolumn{3}{c}{\textbf{Contact}} \\
\midrule
Correct strikes    & \(\displaystyle \sum_{e\in\mathcal{E}_t}\mathbb{I}\{\text{drum}(e)\in D_i\}\) & \(+1.0\) \\
Wrong strikes      & \(\displaystyle \sum_{e\in\mathcal{E}_t}\mathbb{I}\{\text{drum}(e)\notin D_i\}\) & \(-0.5\) \\
Missed strikes     & \(\displaystyle \sum_{d\in D_i}\mathbb{I}\{\text{drum missed} = d\}\) & \(-2.0\) \\
Proximity   & \(\displaystyle \mathbb{I}\{|D_i|>0\}\sum_{s\in\{s_{\text{L}}, s_{\text{R}}\}}min_{d\in D_i}\|p_{s,t}-p_d\|\) & \(-1.0\) \\
\midrule
\multicolumn{3}{c}{\textbf{Regularization}} \\
\midrule
Action‑rate        & \(\displaystyle \|a_t - a_{t-1}\|^2 \) & 
\(-1e^{-3}\) \\
DoF acceleration   & \(\displaystyle \|\ddot{q}\|^2 \)  & 
\(-2.5e^{-7}\) \\
\bottomrule
\end{tabular}}
\begin{tablenotes}
\footnotesize
\item[1] Note: \(\mathcal{E}_t\) is the set of drums the robot actually hits at time \(t\); \(p_{s,t}\) is the position of stick \(s\) at time \(t\); \(p_d\) is the position of drum \(d\).
\end{tablenotes}
\end{table}

\subsubsection{PD Controller}
Table~\ref{tab:pd_param} lists the stiffness ($K_p$) and damping ($K_d$) values used in the PD controller for all joints.

\begin{table}[htbp]
\centering
\caption{PD Controller Parameters}
\label{tab:pd_param}
\begin{tabular}{lcc}
\toprule
Joint &
\makecell{$K_{p}$ \\ (N$\cdot$m/rad)} &
\makecell{$K_{d}$ \\ (N$\cdot$m$\cdot$s/rad)} \\
\midrule
Waist & 100 & 2 \\
Shoulder pitch & 90 & 2 \\
Shoulder roll & 60 & 1 \\
Shoulder yaw & 20 & 0.4 \\
Elbow & 60 & 1 \\
Wrist roll & 20 & 0.1 \\
Wrist pitch & 4 & 0.2 \\
Wrist yaw & 4 & 0.2 \\
\bottomrule
\end{tabular}
\end{table}

\subsubsection{Policy Training and Evaluation Protocol}
We use a 3-layer MLP architecture for both the actor and critic networks, with hidden sizes of (128, 64, 32). For each song or training condition, policies are learned across 5 random seeds over 1500 PPO updates. Experiments are conducted on a server with Nvidia RTX 4090 GPUs, running 8192 parallel environments in Isaac Gym. After training, each policy is evaluated over 20 independent rollouts, and performance is measured using the mean F1 score between actual and target drum strikes across all timesteps. The final performance metric is the average F1 score and standard deviation across the 5 seeds.

\subsection{Additional Results}

\subsubsection{Song-level Performance Analysis}
\begin{figure*}[htbp]
    \centering
    \includegraphics[width=0.9\textwidth]{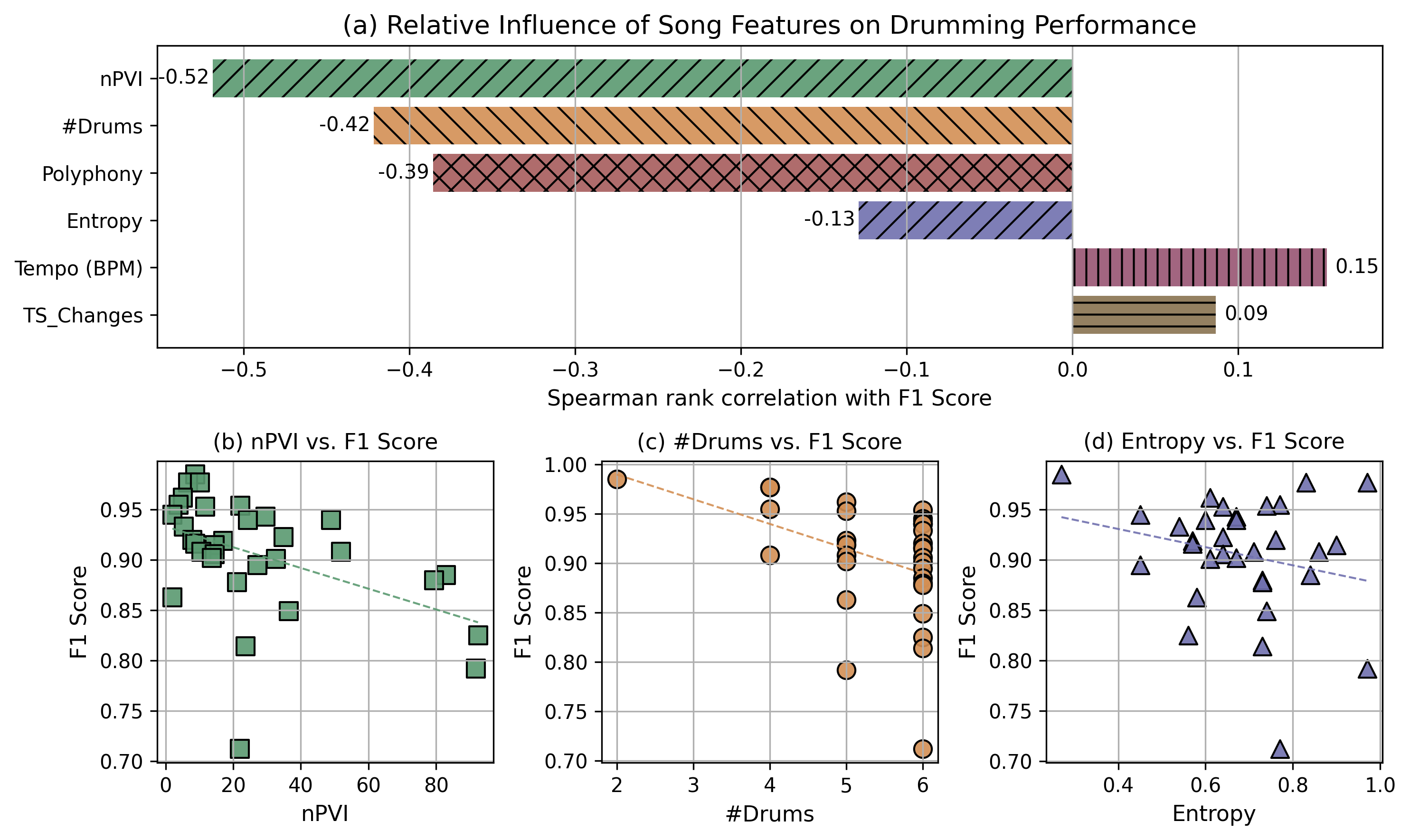}
    \caption{
    Song-level performance analysis across specialist policies. (a) Absolute Spearman correlations between song features and per-song F$_1$. (b–d) Scatter plots showing F$_1$ versus nPVI, number of drums, and entropy.
    }

    \label{fig:song_metrics_f1score}
\end{figure*}

Figure~\ref{fig:song_metrics_f1score} presents song-level performance analysis across all specialist policies. (a) shows absolute Spearman rank correlations between individual song features and per-song F$_1$ score, while (b–d) show the relationship between F$_1$ and selected features using scatter plots with linear trendlines. Together, these plots summarize how performance varies across songs as a function of rhythmic and spatial attributes of the target drum patterns.

\subsubsection{Temporal Decomposition Ablation}
Table~\ref{tab:temp_decom} reports per-song F$_1$ scores (mean $\pm$ standard deviation over five seeds) obtained with and without temporal decomposition. Temporal decomposition trains fixed-length segments concurrently rather than end-to-end on full tracks. Although the policies achieve similar F$_1$ scores in both cases, wall-clock time is substantially reduced for each run with temporal decomposition (from 8–9 hours to 2–3 hours).

\begin{table}[htbp]
\centering
\caption{F1 Scores With and Without Temporal Decomposition}
\label{tab:temp_decom}
\begin{tabular}{lcc}
\toprule
Song & w/o Decomposition & w/ Decomposition \\
\midrule
Rebel Rebel & 0.98 $\pm$ 0.01 & 0.99 $\pm$ 0.00 \\
Lithium & 0.96 $\pm$ 0.02 & 0.95 $\pm$ 0.02 \\
Fire & 0.93 $\pm$ 0.02 & 0.94 $\pm$ 0.02 \\
In The End & 0.92 $\pm$ 0.02 & 0.90 $\pm$ 0.02 \\
Livin' on a Prayer & 0.86 $\pm$ 0.02 & 0.89 $\pm$ 0.03 \\
Roxanne & 0.89 $\pm$ 0.02 & 0.88 $\pm$ 0.02 \\
\bottomrule
\end{tabular}
\end{table}

\subsubsection{Listener Study}
The listener study evaluates perceptual aspects of the drumming pattens performed by Robot Drummer using subjective ratings collected from human participants. Fifteen participants each watched and listened to three performance videos of specialist policies. After watching the videos, participants answered a series of questions, rating the robot's drumming on a 1–5 Likert scale (1~=~not at all; 5~=~very much) for the following aspects:

\begin{itemize}
    \item Timing Consistency - How consistent and well-timed was the rhythm of the drumming?
    \item Expressiveness - To what extent did the robot convey a sense of musical feel?
    \item Naturalness - How natural or human‑like did the robot's drumming sounded?
    \item Enjoyment - Overall enjoyment of the performances.
\end{itemize}

Across all responses, we observed the following results: 

\begin{itemize}
    \item Timing Consistency: $3.69~\pm~0.63$
    \item Expressiveness: $3.38~\pm~0.96$
    \item Naturalness: $2.46~\pm~0.78$
    \item Enjoyment: $3.08~\pm~1.60$
\end{itemize}

These results show that listeners generally perceived the robot's timing and musical feel in the upper mid-range, with lower ratings for human-like naturalness and mixed enjoyment responses reflected by greater variance.

\begin{table*}[t]
\centering
\caption{Quantitative evaluation of policies on songs with non-zero polyphony}
\label{tab:nonzero_poly}
\resizebox{0.96\textwidth}{!}{
\small
\begin{tabular}{l c c c c c c}
\toprule
Song                        & \#Drums & Entropy & nPVI &  BPM  & Polyphony (\%) & F$_1$ $\uparrow$ \\
\midrule
Limp Bizkit - Break Stuff                 & 4 & 0.83 & 10.22 & 107 & 0.27 & 0.977 $\pm$ 0.005 \\
Metallica - Nothing Else Matters          & 5 & 0.61 & 5.08 & 69 & 0.23 & 0.962 $\pm$ 0.003 \\
Foo Fighters - Everlong                   & 4 & 0.77 & 3.79 & 156 & 0.86 & 0.955 $\pm$ 0.017 \\
Metallica - One                           & 5 & 0.64 & 11.74 & 142 & 1.21 & 0.953 $\pm$ 0.006 \\
AC/DC - You Shook Me All Night Long       & 6 & 0.45 & 2.05 & 129 & 1.71 & 0.945 $\pm$ 0.017 \\
Europe - The Final Countdown              & 6 & 0.60 & 48.96 & 118 & 1.56 & 0.940 $\pm$ 0.011 \\
Rage Against the Machine - Killing in the Name  & 6 & 0.67 & 24.38 & 102 & 0.14 & 0.940 $\pm$ 0.006 \\
Eagles - Hotel California                 & 6 & 0.54 & 5.38 & 149 & 0.98 & 0.933  $\pm$ 0.020 \\
Iron Maiden - Where Eagles Dare           & 5 & 0.64 & 34.82 & 120 & 1.16 & 0.923 $\pm$ 0.016 \\
Green Day - Boulevard of Broken Dreams    & 6 & 0.76 & 7.96 & 167 & 0.20 & 0.920 $\pm$ 0.009 \\
Nirvana - The Man Who Sold the World      & 5 & 0.57 & 17.08 & 118 & 0.40 & 0.919 $\pm$ 0.017 \\
Nirvana - Come as You Are                 & 6 & 0.57 & 8.73 & 120 & 0.71 & 0.916 $\pm$ 0.013 \\
Nirvana - Dumb                            & 6 & 0.90 & 14.59 & 120 & 1.60 & 0.915 $\pm$ 0.050 \\
Metallica - Enter Sandman                 & 5 & 0.86 & 10.63 & 123 & 0.55 & 0.908 $\pm$ 0.025 \\
The Police - Message in a Bottle          & 6 & 0.64 & 14.55 & 150 & 0.21 & 0.906 $\pm$ 0.021 \\
Guns N’ Roses - Sweet Child o’ Mine       & 5 & 0.67 & 13.66 & 128 & 0.21 & 0.902 $\pm$ 0.040 \\
AC/DC - Thunderstruck                     & 6 & 0.45 & 27.17 & 107 & 3.36 & 0.895 $\pm$ 0.013 \\
Guns N’ Roses - Knockin' on Heaven's Door & 6 & 0.73 & 79.42 & 128 & 0.31 & 0.880 $\pm$ 0.023 \\
Iron Maiden - Wasted Years                & 5 & 0.58 & 2.02 & 152 & 4.58 & 0.863 $\pm$ 0.013  \\
Bon Jovi - Bed of Roses                   & 6 & 0.74 & 36.49 & 60 & 0.64 & 0.849 $\pm$ 0.011  \\
Bon Jovi - It's My Life                   & 6 & 0.56 & 92.40 & 120 & 1.69 & 0.825 $\pm$ 0.022 \\
Bon Jovi - Always                         & 6 & 0.73 & 23.61 & 67 & 3.28 & 0.814 $\pm$ 0.020 \\
Rolling Stones - Paint It Black           & 5 & 0.97 & 91.76 & 110 & 3.20 & 0.792 $\pm$ 0.055 \\
Nirvana - Smells Like Teen Spirit         & 6 & 0.77 & 22.02 & 119 & 14.16 & 0.712 $\pm$ 0.030 \\
\bottomrule
\end{tabular}}
\begin{tablenotes}
\footnotesize
\item[1] All songs have zero time signature changes, except for "Metallica - Nothing Else Matters" (17), and  "Nirvana - The Man Who Sold the World" (4).
\end{tablenotes}
\end{table*}

\end{document}